\documentclass[10pt,twocolumn,letterpaper]{article}

\usepackage[pagenumbers]{cvpr} 

\usepackage{graphicx}
\usepackage{amsmath}
\usepackage{amssymb}
\usepackage{booktabs}

\usepackage{amsmath}
\usepackage{multirow}
\usepackage[table]{xcolor}
\usepackage{array}
\usepackage{makecell}

\usepackage{pifont}
\newcommand{\cmark}{\ding{51}}%

\newcommand\blfootnote[1]{%
  \begingroup
  \renewcommand\thefootnote{}\footnote{#1}%
  \addtocounter{footnote}{-1}%
  \endgroup
}

\usepackage[pagebackref,breaklinks,colorlinks]{hyperref}

\usepackage[capitalize]{cleveref}
\crefname{section}{Sec.}{Secs.}
\Crefname{section}{Section}{Sections}
\Crefname{table}{Table}{Tables}
\crefname{table}{Tab.}{Tabs.}

\def\cvprPaperID{2928} 
\def\confName{CVPR}
\def\confYear{2023}

\begin{document}

\title{Weakly Supervised Monocular 3D Object Detection using 
\\ Multi-View Projection and Direction Consistency}

\author{
Runzhou Tao$^{1,3*}$,
Wencheng Han$^{2*}$,
Zhongying Qiu$^3$,
Cheng-zhong Xu$^{2}$,
Jianbing Shen$^{2\dagger}$\\
$^1$ Beijing Institute of Technology,
$^2$ SKL-IOTSC, CIS, University of Macau,
$^3$ QCraft \\
{\tt\small \{wencheng256, shenjiangbingcg\}@gmail.com} \\
 \href{https://github.com/weakmono3d/weakmono3d}{https://github.com/weakmono3d/weakmono3d}
}

\maketitle

\begin{abstract}
{Monocular 3D object detection has become a mainstream approach in automatic driving for its easy application.}
A prominent advantage is that it does not need LiDAR point clouds during the inference. However, most current methods still rely on 3D point cloud data for labeling the ground truths used in the training phase. 
{This inconsistency between the training and inference makes it hard to utilize the  large-scale feedback data and increases the data collection expenses.}
To bridge this gap, we propose a new weakly supervised monocular 3D objection detection method, which can train the model with only 2D labels marked on images. To be specific, we explore three types of consistency in this task, \ie the projection, multi-view and direction consistency, and design a weakly-supervised architecture based on these consistencies. 
{Moreover, we propose a new 2D direction labeling method in this task to guide the model for accurate rotation direction prediction.} Experiments show that our weakly-supervised method achieves comparable performance with some fully supervised methods. When used as a pre-training method, our model can significantly outperform the corresponding fully-supervised baseline with only 1/3 3D labels.

\end{abstract}

\blfootnote{$*$Equal contribution. $\dagger$Corresponding author: \textit{Jianbing Shen}.
This work was supported in part by the FDCT grant SKL-IOTSC(UM)-2021-2023, the FDCT Grant 0123/2022/AFJ, the Grant MYRG-CRG2022-00013-IOTSC-ICI, and the Grant SRG2022-00023-IOTSC.
%
}

\section{Introduction}
\label{sec:intro}
Monocular 3D object detection is a foundational research area in computer vision and plays an important role in autonomous driving systems. It aims to identify the objects and estimate the 3D bounding boxes of the corresponding targets with a single image as input. Different from 3D point clouds detection methods like~\cite{shi2019pointrcnn,zarzar2019pointrgcn,vora2020pointpainting,shin2019roarnet,wang2019frustum}, monocular 3D detection models~\cite{zou2021devil,wang2021progressive,wang2021depth,dd3d,ma2020rethinking} alleviate the need of LiDAR sensors, making the self-driving system easier to be applied. 

\begin{figure}
  \centering
    \mbox{}\hfill
  \includegraphics[width = 0.99 \linewidth]{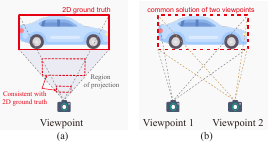}
  \hfill\mbox{}  \mbox{}\hfill
  \vspace{-6mm}
  \caption{\textbf{Illustration of the projection and multi-view consistency.} (a) Only projection consistency cannot determine the accurate position of the target because projection loss has more than one optimal solution in the 3D space. 
  {For example, the two dashed boxes in 3D space produce the same projection loss because they have the same projection in 2D space.} 
  (b) Constrained by the multi-view consistency, the optimal solution must be the common solution for two viewpoints, that is, the target location.
  }
  \label{Fig:part12}
\vspace{-5mm}
\end{figure} 

However, there is still a challenging problem that limits the application of 3D object detection with pure camera vision data. That is, the ground truth 3D boxes used in the training phases are usually labeled with 3D point clouds~\cite{kitti,waymo,nuscene}. Recently, self-driving systems with pure camera vision inputs have become a new trend. 
{But the feedback video clips captured by the production cars cannot be utilized to improve the 3D object detection models because of the lack of training labels.} 
{Compared with the data from the data-collection cars, the feedback images from production cars have a larger scale diversity and contain more corner cases, which are crucial for improving the robustness of models.} 

In this paper, we propose a new weakly supervised training method, which can train the 3D object detection models with only camera images and 2D labels, making it possible to utilize the feedback data from the production cars. To achieve this goal, we exploit three types of consistency between the 3D boxes and 2D images and fully utilize them for training the object detection models. The first is the projection consistency. With the intrinsic matrix of a camera, a 3D box predicted by the models can be projected into the 2D image space, and the projected boxes should be consistent with the corresponding 2D boxes. 
Based on this, we propose a projection loss by minimizing the difference between the projected boxes and 2D ground truths. This criterion can guide the predicted 3D boxes into the projection regions. However, only projection consistency cannot provide enough information to correct the errors in the 3D space, especially for the depth dimension, as shown in Fig.~\ref{Fig:part12} (a). According to the perspective principle, multiple boxes in the 3D space can be projected into the same 2D box in the image. Thus, there is more than one optimal solution for the projection loss, and the errors caused by these boxes cannot be optimized by the projection loss.

Aiming at solving this limitation, we incorporate multi-view consistency into our method to minimize the errors in 3D space. The same object captured from different viewpoints would show different positions and shapes in the corresponding 2D images. But in the 3D space, 3D bounding boxes belonging to the same object should be consistent, \ie they should be of the same position, size and rotation angle in a certain coordinate system. 
Based on this, we construct the multi-view consistency by minimizing the discrepancy between the predicted bounding boxes of the same object from a different point of view. As shown in Fig.~\ref{Fig:part12}(b), projection losses on the two viewpoints will constrain the predictions into their projection regions, and the multi-view consistency will further guide the predictions to the common optimal solutions of the two views, which are where the objects located. Notably, in our work, images paired from different viewpoints are only used for calculating the losses, and the models {still only take monocular inputs} in the evaluation phase.

The last consistency presented in this paper is the direction consistency for guiding the prediction of the direction scale. 
{In previous works \cite{nuscene} \cite{waymo} \cite{kitti}, 3D rotation direction is labeled on point clouds by a vector from the center to the front of objects.}
To avoid the need for 3D LiDAR data, we propose a new labeling method named 2D direction label directly on pure camera images, indicating the 2D direction of the object in the images. The predicted 3D box rotation should be consistent with the direction in 2D space when they are projected, \ie the direction consistency.
Based on this consistency, we further design a 2D rotation loss for optimizing the rotation-scale estimation.

The proposed weakly supervision method is a general framework that can be integrated with most monocular 3D detection models.
To show the efficiency of our method, we incorporate it with {a representative model - DD3D~\cite{dd3d} in this task}, and evaluate it on the KITTI benchmark. 
Results show that our method can achieve comparable performance with some fully supervised methods. Also, to demonstrate the application in real scenes, we collect a new dataset named \textbf{ProdCars} from production cars and evaluate the performance of our method on it.

In short, our contributions are summarized in four folds:
\begin{itemize}
    \item We propose a new weakly supervised method for monocular 3D object detection, which only incorporates 2D labels as ground truth without needing any 3D point clouds for labeling. To the best of our knowledge, we are the first work that totally {avoid the dependency} of 3D point clouds in this task.
    \item We incorporate projection consistency and multi-view consistency into this task and design two consistency losses for guiding the prediction of the accurate 3D bounding boxes based on them.
    \item {We propose a new type of labeling method named 2D direction label for replacing the 3D rotation label marked on the point clouds data}, as well as a direction consistency loss based on the new labels.
    \item In our experiments, the proposed weakly supervised method achieves comparable performance with some fully supervised methods. We also fine-tune our model with a small proportion of 3D ground truths. Results show that even with only 1/3 of the ground truth labels, our method can achieve better performance than the corresponding fully supervised baselines, {showing the potential for improving models based on feedback production data.}
\end{itemize}

\section{Related Works}
\subsection{Monocular 3D Object Detection}
Monocular 3D object detection is a challenging task because of the lack of object information in 3D space. Previously, lots of monocular 3D object detection methods~\cite{chen2016monocular,mousavian20173d,qin2019monogrnet,wang2021fcos3d,liu2019deep,liu2020smoke} aim to find the clues benefited for 3D object detection from 2D images. 
{Deep3DBox~\cite{mousavian20173d} detected 2D object MS-CNN~\cite{cai2016unified} by 2D object detector and estimated the dimension and orientation of the object by a 3D regression head. Afterwards, the network combines geometric projection constraints to obtain a 3D pose. FCOS3D~\cite{wang2021fcos3d} is an improved 3D objetc detection method based on FCOS~\cite{tian2019fcos}, which decoupled the 7-DoF 3D object into 2D and 3D attributes, and assigns an object to different feature levels by the 2D dimension of the object.
%
%
DD3D~\cite{dd3d} noticed that the depth information is highly critical to the monocular 3D object detection task. It learned the dense depth estimation in advance, and significantly improved the 3D object detection performance.
According to the geometric relationship of 3D objects, MonoDDE~\cite{li2022diversity} generated multiple estimations for each object, and formulated the depth selection with combination strategies to make the final depth more accurate.} 

Recently, more well-known algorithms~\cite{li2020rtm3d,li2021monocular,chen2020monopair} have been proposed to explore the useful information in 2D space to further constrain the 3D detection results. 
Based on the original 3D regression head, MonoCon~\cite{liu2022learning} added some additional auxiliary tasks to predict the object center point and corner point in the 2D image to better learn the correction information. MonoJSG~\cite{lian2022monojsg} formulated the depth estimation as a progressive refinement procedure and proposed a joint semantic and geometric cost volume to evaluate the depth error. It obtained the depth of the object more accurately by segmenting the object and estimating the depth of the dense points in the object.
In addition, to directly feed monocular images into the network, ~\cite{wang2019pseudo,you2019pseudo} used the depth estimation network~\cite{fu2018deep,chang2018pyramid} to predict the depth value of each pixel for the monocular image, and represented the image in the form of the 3D point cloud. 
{Thus, the 3D object detector can be used to detect it; however, due to the disadvantage of high delay, {it is hard to meet the real-time requirement of automatic driving scenes.}}

\subsection{LiDAR 3D Object Detection}
{Compared with images, LiDAR point clouds have disorder characteristics, rotation invariance and uneven distribution, which leads to the 2D convolutional network cannot be directly applied for 3D point cloud data.} 
Therefore, it is particularly important to extract information from the 3D point cloud data directly~\cite{wang2022ssda3d,li2022lwsis, yin2021graph,meng2021towards,yin2022semi,yin2022proposalcontrast,yin2020lidar}.
{PointNet~\cite{qi2017pointnet} directly extracted high-level features from the coordinate of the 3D point clouds through MLP, and aggregated the global features by max-pooling layers to solve the disorder problem of the point clouds data.}
{Based on PointNet~\cite{qi2017pointnet}, PointNet++~\cite{qi2017pointnet++} recursively aggregated the 3D point clouds in a layered manner and proposed a multi-scale and multi-resolution grouping scheme. 
%
%
Unlike the methods of directly using 3D point clouds, VoxelNet~\cite{zhou2018voxelnet} divided the 3D point clouds into equidistant 3D voxels and encoded each voxel to output the 3D object detection result through the region proposal network. 
MV3D~\cite{chen2017multi} projected the 3D point clouds into a 2D bird's-eye grid, and set a large number of 3D anchor boxes in advance to generate the 3D bounding box.
%
%
PointPillar~\cite{lang2019pointpillars} encoded the 3D point clouds into a vertical column as a special voxel. 
Then, the 3D point clouds in the cylinder are encoded through PointNet~\cite{qi2017pointnet} to generate a pseudo-2D image and finally use the 2D object detector to detect 3D objects.
}

\subsection{Weakly Supervised 3D Object Detection}
{For 3D object detection task, data annotations for 3D point clouds are very expensive, and so some works explore how to use more straightforward annotations to train 3D object detection task. WS3D~\cite{meng2020weakly} presented a weakly supervised method for 3D LiDAR object detection with two stages. 
In the first stage, the cylindrical object proposal is generated by clicking annotations in the bird's-eye-view. Afterwards, the network uses a few precisely labeled object instances to optimize the cylindrical proposal to generate the final 3D object bounding box.
VS3D~\cite{qin2020weakly} learned from the point cloud to generate the 3D proposal and uses a 2D classification network to identify the 3D proposal. 
WeakM3D~\cite{peng2022weakm3d} first detected the image and combined the 3D point clouds with the detection results to obtain the object-LiDAR-points. 
In addition, it proposed a method to estimate object orientation $\theta$, which obtained the orientation of each pair of points in the object-LiDAR-points. 
}
{On the contrary, our method not only does not rely on the 3D point clouds data, but also achieve comparable results with some fully-supervised methods.}

\section{The Proposed Method}

\begin{figure*}[t]
  \centering
  \resizebox{\textwidth}{!}{
  \includegraphics{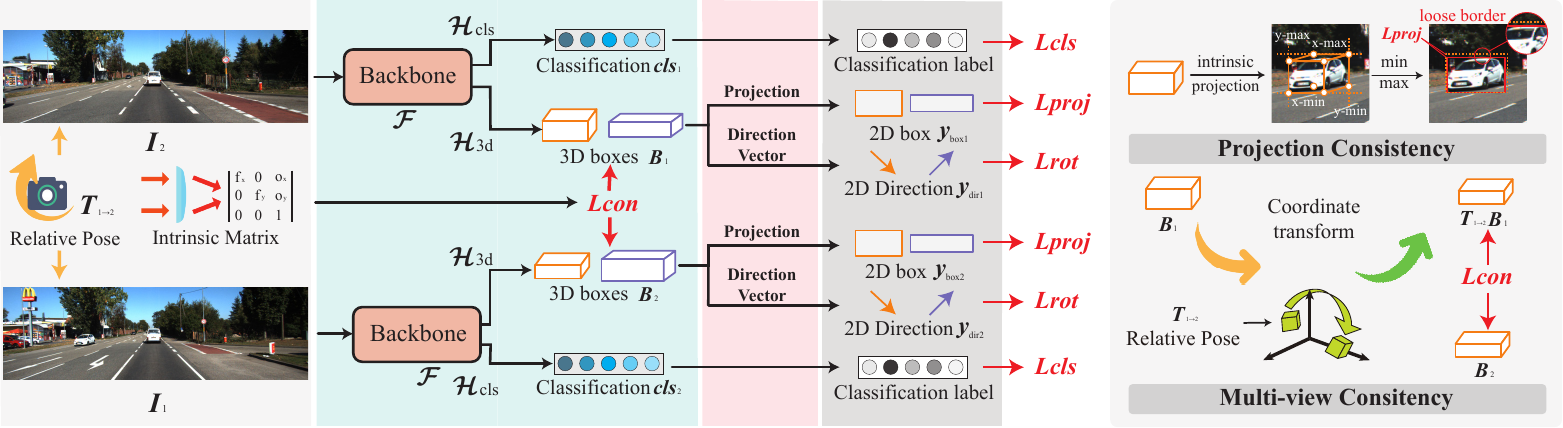}}
  \vspace{-2mm}
  \caption{\textbf{The architecture of the proposed method.} The left column shows that during the training phase, image pairs from different viewpoints are sent into the detection model, and 4 losses are computed between the predictions and the 2D ground truth.
  The right column shows the details of the projection consistency and the multi-view consistency. 
  {To calculate the projection consistency loss, we project the predicted box into the 2D image and convert it into a 2D box and finally compute the differences between the 2D box and 2D box labels.}
  To calculate the consistency loss, we first transform the predicted 3D boxes from viewpoint 1 into the coordinate system of viewpoint 2 and then compute the difference between the converted boxes and the predicted boxes of viewpoint 2.
  }
  \label{Fig:architecture}
\end{figure*} 

\subsection{Architecture}
\label{Sec:network} 
Fig.~\ref{Fig:architecture} shows an overview of the proposed architecture, in which the proposed method is general and can be incorporated with most previous monocular 3D detection methods.
{To show the efficiency, we employ the representative DD3D~\cite{dd3d} as our base detection model, keeping all the settings the same as our baseline except for the training method.
}
In the training phase, image pairs taken from two different viewpoints and their relative positions are needed. According to the available data, image pairs can be retrieved from different sources. For data from multiple cameras, the pairs could be the images from different cameras with some overlap areas. For data captured by only a single camera, the pairs could be adjacent frames from video sequences when the camera is moving, and the relative positions could be predicted by a separated network like~\cite{zhou2017unsupervised}. Firstly, the image pairs ($\boldsymbol{I_1}$, $\boldsymbol{I_2}$) are sent into the backbone model $\mathcal{F}$ for feature extraction and produce the corresponding feature maps $\boldsymbol{f_1}$, $\boldsymbol{f_2}$:
\begin{equation}
    \boldsymbol{f_1} = \mathcal{F}(\boldsymbol{I_1}, W); \boldsymbol{f_2} = \mathcal{F}(\boldsymbol{I_2}, W),
\end{equation}
where $W$ is the weight of the backbone model. Then, different heads are employed to generate predictions for the objects from the feature maps $\boldsymbol{f_1}$, $\boldsymbol{f_2}$:
\begin{equation}
    cls_i = \mathcal{H}_{cls}(\boldsymbol{\boldsymbol{f_i} }); B_i = \mathcal{H}_{3d}(\boldsymbol{\boldsymbol{f}_i}), i \in \{1, 2\},
\end{equation}
where $\mathcal{H}_{cls}$ is the classification head and $\mathcal{H}_{3d}$ is the 3D detection heads. $B$ is the predicted 3D boxes for the targets. 

The predictions and 2D labels are used to calculate the losses. The projection loss $L_{proj}$ and direction consistency loss $L_{rot}$ are applied on the results of each image, and the multi-view consistency loss $L_{con}$ is calculated between the results of the pairs:
\begin{equation}
    \begin{aligned}
    \mathcal{L}_{proj} &= E_{proj}(B_1, y_{box1}) + E_{proj}(B_2, y_{box2}); \\
    \mathcal{L}_{rot} &= E_{rot}(B_1, y_{dir1}) + E_{rot}(B_2, y_{dir2}); \\
    \mathcal{L}_{con} &= E_{con}(B_1, B_2)
    \end{aligned}
\end{equation}
where {$y_{box}$ denotes the 2D boxes labels on the image and $y_{dir}$ represents the 2D direction labels} introduced in \S~\ref{Sec:rotation}. $E$ are the criterion of corresponding losses.
Finally, the three losses and classification loss $\mathcal{L}_{cls}$ are summed together to be the final loss $\boldsymbol{\mathcal{L}}$:
\begin{equation}
    \boldsymbol{\mathcal{L}} = \mathcal{L}_{proj} +  \mathcal{L}_{con} + \mathcal{L}_{rot} + \mathcal{L}_{cls}.
\end{equation}

{During the inference, the model only takes one image from a single viewpoint and produces the 3D boxes for the targets.} In the rest of this section, we will introduce three consistencies between the 2D labels and 3D predictions and how we use them for guiding the optimization of models.

\subsection{Projection Consistency}
\label{Sec:projection_loss}
{To avoid the need of the 3D point clouds annotations in the labeling procedure, we only employ 2D ground truths of images in our work.}
An intuitive idea is that the predicted 3D boxes can be projected into the 2D image space, and the projected 2D boxes should be consistent with the labeled 2D boxes. We define this property as the projection consistency.

Given the intrinsic matrix of a camera, a point $\boldsymbol{Q}(x, y, z)$ in the real world 3D space can be converted into a point $\boldsymbol{q}$ in the 2D image space by multiplying with the matrix:
\begin{equation}
   z \boldsymbol{q} = \left[ \begin{array}{ccc}
        f_x & 0 & o_x \\
        0 & f_y & o_y\\
        0 & 0 & 1
        \end{array} 
        \right ] \boldsymbol{Q} 
\end{equation} 
where $f_x$ and $f_y$ are the pixel focal lengths and $o_x, o_y$ are the offsets of the principal point. Therefore, the 3D boxes $B$ with 8 corner points can be transformed into the corresponding points in the 2D space. Then a 2D bounding box $b$ can be generated by calculating the minimum and maximum values of the x,y coordinates, as shown in Fig.~\ref{Fig:architecture}. Finally, the loss values are computed between the $b$ and the 2D ground truth $y_{box}$. 

When estimating the error, there are some significant differences between our method and the traditional 2D box loss. {The first is that there exists a gap between the 2D loss and the 3D evaluation metrics.} 
We only compute the 2D box difference in the training phase, but the 3D box difference is the target we want to minimize in the evaluation. The 3D box of an object is usually translation invariant, \ie no matter whether the object is near to the camera or far from it, the 3D box is still the same. But the 2D box will change according to the perspective principle. 
This gap would introduce some undesired problems. For example, objects far from the camera have small bounding boxes in the 2D space, but their evaluation is conducted on the original size. That means even a small difference between the 2D boxes would cause large errors in the 3D space. 
To balance the performance of near and far objects, we employ a size-independent $GIoU$ loss~\cite{rezatofighi2019generalized} as our main criterion. As the discrepancy of near boxes is enlarged by the perspective, we employ an extra $L_1$ loss to improve the detection performance on near targets. 

Secondly, since the targets are not exact cubes, the projected boxes cannot perfectly align with the 2D boxes. 
{As shown in Fig.~\ref{Fig:architecture}, the converted boxes will introduce the loose boundaries on the images, while the labeled 2D boxes do not.} 
Considering this, we replace the traditional $L_1$ by SmoothL1~\cite{girshick2015fast} in our loss function, and leave a soft margin when computing the difference between the converted boxes and 2D ground truth.
If the difference between the boxes is smaller than a given threshold, the gradient will be reduced as:
\begin{equation}
    SmoothL_1 = \left\{  \begin{array}{ll}
        	\left| a - b \right| - 0.5 \times \gamma ,&  if  \left| a - b \right| > \gamma \\
        0.5(a - b)^2 / \gamma,&  otherwise 
    \end{array} 
\right., 
\end{equation}
where $a,b$ are the input boxes and $\gamma$ is the soft margin.
Finally, {the projection loss is defined as}:
\begin{equation}
    E_{proj} =  GIoU(b, y_{box}) + \lambda SmoothL_1(b, y_{box}, \gamma) ,
\end{equation}
where $\lambda$ is the balance ratio.

\subsection{Multi-View consistency}
\label{Sec:consistency_loss}
The projection consistency would constrain the predicted 3D boxes into the corresponding projection region, where all the 3D boxes have the same 2D projection box, as shown in Fig.~\ref{Fig:part12} (a). This means that there is more than one optimal model solution with only the projection loss. 
{However, the real and exact position of the target is the only solution we want to get, so we need to find more constrains to achieve this goal.}

{If we observe an object from two different viewpoints at the same time, an object will show different projection views in the two images, but its 3D bounding box in the real world is the same.} 
Therefore, there exists an inner consistency between the observations from different viewpoints, and we call this multi-view consistency. As shown in Fig.~\ref{Fig:part12}(b), two different viewpoints have two independent projection regions, each of which has a set of optimal solutions constrained by the projection loss. But when fixing the rotation angle of objects, there is only one common solution of the two sets, which is the target box. Besides the position, all the 3D attributes of one target would be the same when observed from a different point of view, such as the size and the direction.

To utilize this consistency, we should first convert the 3D boxes from different viewpoints into the same coordinate system. So we need a relative position matrix $\boldsymbol{T}_{1\rightarrow2}$ between the viewpoints. There are several ways to get this information. For data captured by multiple cameras, the relative position is fixed after the cameras are installed. For video sequences captured by a single camera, the relative position can be predicted by a PoseNet~\cite{zhou2017unsupervised} trained by the self-supervised methods. 

\begin{figure}
  \centering
 \mbox{}\hfill
 \includegraphics[width = 0.99 \linewidth]{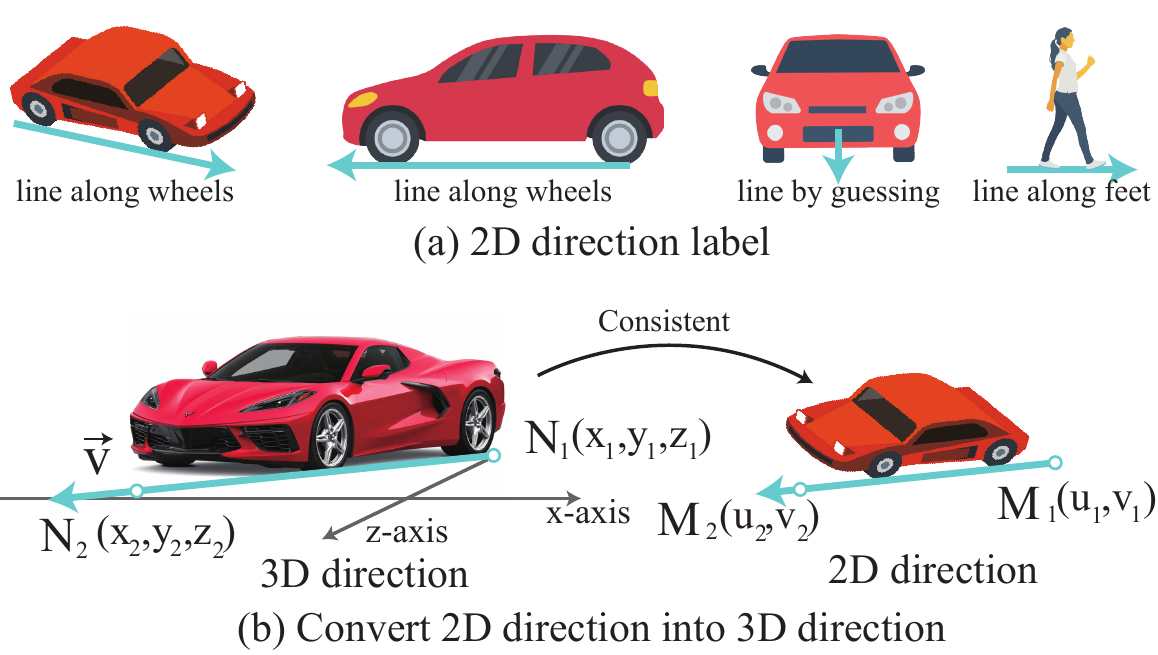}
  \hfill\mbox{}  \mbox{}\hfill
  \caption{\textbf{Illustration of the Direction Consistency.} (a) Examples of how the 2D directions are labeled. For most circumstances, we can mark the 2D direction by drawing a parallel line of the wheels or feet. (b) Given a 2D direction, we can convert it into the corresponding 3D direction according to consistency.
  }
  \label{Fig:rotation}
\vspace{-5mm}
\end{figure} 

Given the 3D boxes $B_1$ and $B_2$ from two different viewpoints, we first convert $B_1$ from its coordinate system into that of $B_2$. Then we calculate the difference between the converted box and $B_2$. In this module, we employ an $L_1$ loss as the criterion:
 \begin{equation}
E_{con} =  L_1(B_2, \boldsymbol{T}_{1\rightarrow2} B_1)
\end{equation}

There is stationary a problem when the viewpoint data is extracted from video sequences. That is, some of the objects are not still when capturing the two frames, which would cause inconsistency between the two viewpoints. 
{To solve this problem, we employ a simple but efficient strategy by labeling the objects as inconsistent if they move obviously in the frame, without computing the corresponding multi-view losses produced by them.}

\begin{figure*}[t]
  \centering
  \resizebox{\textwidth}{!}{
  \includegraphics{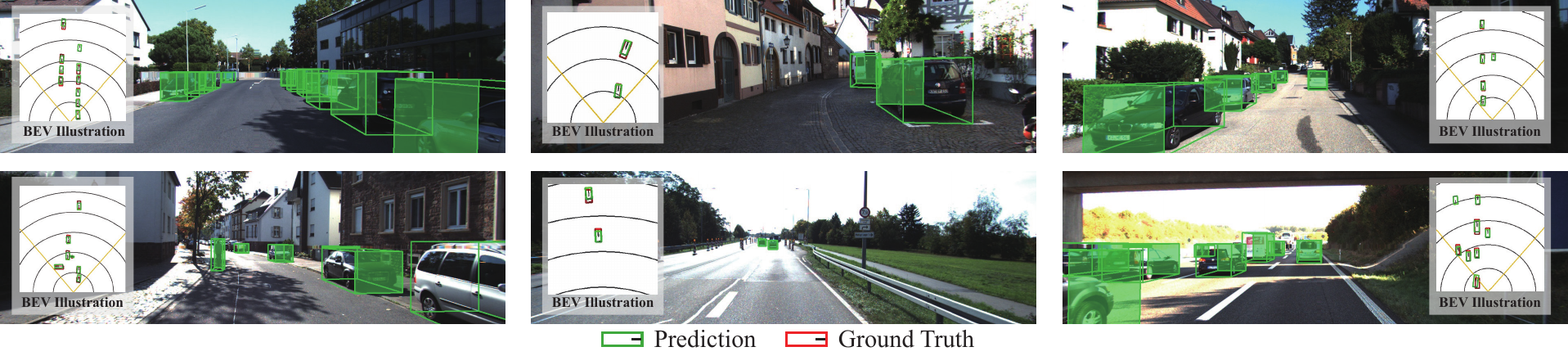}}
  \vspace{-5mm}
  \caption{\textbf{Visualizations of our WeakMono3D results.} Both the solid side in the image and the short bar in the BEV-box indicate the direction of the object.}
  \label{Fig:illustration}
\vspace{-5mm}
\end{figure*}

\subsection{Direction Consistency}
\label{Sec:rotation}
In most previous approaches~{\cite{kitti,nuscene,waymo}}, the rotation angels are labeled with 3D point clouds. Without 3D information, it is quite difficult to indicate accurate rotation angles for humans. In this paper, instead of labeling the rotation angles, we propose a new method to get the rotation according to the direction consistency. 

If we draw a vector in the 3D space to indicate the direction of the object, we will get a 2D vector in the image when projecting it into the 2D space. 
{We call 2D vector as 2D direction, and the relationship between it and 3D rotation is called the direction consistency.
The 2D directions are easier to be labeled on the images by humans.
} 
 Fig.~\ref{Fig:rotation}(a) shows some representative examples of the 2D directions. For most of the objects, the 2D direction can be labeled by drawing a line along the wheels on one side of the car or the feet of people. 
 {Even if the wheels are occluded, we can also guess the direction according to the surrounding environment, such as the lane lines.}

Then, we design a new method to recover 3D rotations from the 2D labels. Given a 2D direction vector $\overrightarrow {M_1M_2}$, $M_1=(u_1, v_1)$ and $M_2=(u_2, v_2)$ are the start and end point of the line. 
There are two corresponding points in the 3D spaces named $N_1 = (x_1, y_1, z_1)$ and $ N_2 = (x_2, y_2, z_2)$,
and the 3D direction vector is $\overrightarrow {N_1N_2}$, as shown in Fig.~\ref{Fig:rotation}.  With the intrinsic matrix, the relationship between $M_1, M_2$ and $N_1, N_2$ is defined as:
\begin{equation}
    \left[ \begin{array}{c}
        u_i \\
        v_i \\
        1
        \end{array} 
        \right ]   = \left[ \begin{array}{ccc}
        f_x & 0 & o_x\\
        0 & f_y & o_y\\
        0 & 0 & 1
        \end{array} 
        \right ] \left[ \begin{array}{c}
        x_i/z_i \\
        y_i/z_i \\
        1
        \end{array} 
        \right ] 
\end{equation}
Then, it can be simplified as:
\begin{equation}
    u_i = \frac{x_if_x}{z_i} + o_x; v_i = \frac{y_if_y}{z_i} + o_y, i \in \{1, 2\}.
\end{equation}
{We can further represent $x_i$ and $z_i$ using $u_i$, $v_i$ and $y_i$ as:}
\begin{equation}
    x_i = \frac{y_if_y(u_i - o_x)}{f_x(v_i - o_y)}; z_i = \frac{y_if_y}{v_i - o_y}, i \in \{1, 2\}.
    \label{Eq:x_z}
\end{equation}

For most cases, the targets are in the same plane as the camera. 
So we mainly focus on the rotation in the XZ-plane and presume the direction vector in 3D space is parallel to the ground. 
{Therefore, we assume that all the $y$ will have the same values in Equation~(\ref{Eq:x_z}).} 
Then, the direction vector in the 3D space can be calculated as:
\begin{equation}
    \begin{aligned}
    \vec n &= \boldsymbol{N}_2 - \boldsymbol{N}_1 \\
    &= (x_2 - x_1, z_2 - z_1).
    \end{aligned}
\end{equation}
Finally, the rotation loss can be calculated between the recovered vector and the predicted rotation vector:
\begin{equation}
    E_{rot} = 1-\frac{\vec n \cdot \vec n_p}{|\vec n||\vec n_p|}
    \label{Eq:E_rot}
\end{equation}
where $\vec n_p$ is the predicted rotation vector. In Equation~(\ref{Eq:E_rot}), $y$ is eliminated. Thus, the value of $y$ will not influence the result, and the loss value is determined by $u$ and $v$ values. 

\section{Experiment}
\subsection{Experiment Setup}
\noindent\textbf{Implementation Details.} Our experiments are based on DD3D~\cite{dd3d}, which employs a DLA-34~\cite{yu2018deep} and FPN~\cite{lin2017feature} as the detector’s backbone and neck. We maintain the structure of the original network and keep all hyper-parameters the same as our baseline. The network is trained with a batch size of 8 on 8 NVIDIA Tesla V100 GPUs for 200,000 iterations. The learning rate is set as $2 \times 10^{-3}$, dropped by multiplying 0.1 both at 170,000 and 190,000 iterations. We adopt the SGD optimizer for training. During inference, we save the top 100 detected 3D bounding boxes and use NMS to remove the redundant predictions.

\noindent\textbf{Data Augmentation.} We adopt three data augmentation techniques: random horizontal flip, random resize and random colour jitter. In our implementation, we set a $50\%$ chance to flip the image horizontally, and according to the preset sizes, the picture is randomly resized with an equal ratio. Colour jitter is carried out on three types of transforms: brightness, contrast and saturation, and the proportion of jitter is from $-0.2$ to $0.2$.

\begin{table*}[t]
\centering
\small
\setlength{\tabcolsep}{5.5mm}{
    \begin{tabular}{l|c|c|ccc} 
    \hline
    \multirow{2}*{Method}  & 
    \multirow{2}*{Supervision} & 
    \multirow{2}*{LiDAR} & 
    \multicolumn{3}{c}{AP$_{BEV}$/AP$_{3D}$ (IoU=0.5)$|\scriptstyle R_{40}$} \\ 
    \cline{4-6} & & & Easy & Moderate & Hard \\
    \hline
    CenterNet~\cite{zhou2019objects} & \multirow{6} * {Full} & \multirow{6} * {Label} & 34.36/20.00 & 27.91/17.50 & 24.65/15.57 \\
    MonoGRNet~\cite{qin2019monogrnet} & & & \cellcolor{gray!10} 52.13/47.59 &\cellcolor{gray!10} 35.99/32.28 &\cellcolor{gray!10} 28.72/25.50 \\
    M3D-RPN~\cite{brazil2019m3d} & & & 53.35/48.53 & 39.60/35.94 & 31.76/28.59 \\ 
    MonoPair~\cite{chen2020monopair} & & & \cellcolor{gray!10}61.06/55.38 &\cellcolor{gray!10} {47.63/42.39} &\cellcolor{gray!10} {41.92/37.99} \\
    MonoDLE~\cite{ma2021delving} & & & 60.73/55.41 & 46.87/43.42 & 41.89/37.81 \\
    GUPNet~\cite{lu2021geometry} & & & \cellcolor{gray!10}{61.78/57.62} &\cellcolor{gray!10} 47.06/42.33 &\cellcolor{gray!10} 40.88/37.59 \\
    \hline
    \hline
    VS3D~\cite{qin2020weakly} & \multirow{4} * {Weak} & Train+Val &31.59/22.62 & 20.59/14.43 & 16.28/10.91 \\
    Autolabels~\cite{zakharov2020autolabeling} & &Train  &\cellcolor{gray!10} 50.51/38.31 &\cellcolor{gray!10}  30.97/19.90 &\cellcolor{gray!10}  23.72/14.83\\
    WeakM3D~\cite{peng2022weakm3d} & & Train & 58.20/50.16 & 38.02/29.94 & 30.17/23.11 \\
    Our WeakMono3D & & None & \cellcolor{green!10} 54.32/49.37 & \cellcolor{green!10} 42.83/39.01 & \cellcolor{green!10} 40.07/36.34 \\
    \hline
    \end{tabular}}
    \caption{\textbf{Comparison on KITTI validation set for the car category.} For all results, we use AP$|\scriptstyle R_{40}$ metrics with IoU threshold$=0.5$.LiDAR means where the method uses LiDAR data. We highlight our results in \colorbox{green!10}{green}.}
  \label{tab:kitti_val}
   \vspace{-2mm}
\end{table*}

\begin{table*}
  \small
  \centering
    \setlength{\tabcolsep}{6mm}{
    \begin{tabular}{l|c|c|ccc} 
    \hline
    \multirow{2}*{Method}  & 
    \multirow{2}*{Supervision} & 
    \multirow{2}*{LiDAR} & 
    \multicolumn{3}{c}{AP$_{BEV}$/AP$_{3D}$ (IoU=0.7)$|\scriptstyle R_{40}$} \\ \cline{4-6} & & & Easy & Moderate & Hard \\
    \hline
    FQNet~\cite{liu2019deep} & \multirow{7} * {Full} & \multirow{7} * {Label} & 5.40/2.77 & 3.23/1.51 & 2.46/1.01 \\
    GS3D~\cite{li2019gs3d} &  &  & \cellcolor{gray!10} 8.41/4.47 & \cellcolor{gray!10} 6.08/2.90 & \cellcolor{gray!10} 4.94/2.47 \\
    ROI-10D~\cite{manhardt2019roi} & & & 9.78/4.32 & 4.91/2.02 & 3.74/1.46 \\
    MonoGRNet~\cite{qin2019monogrnet} & & & \cellcolor{gray!10} 18.19/9.61 & \cellcolor{gray!10} 11.17/5.74 & \cellcolor{gray!10} 8.73/4.25 \\
    MonoPSR~\cite{ku2019monocular} & & & \textbf{18.33/10.76} & \textbf{12.58/7.25} & \textbf{9.91/5.85} \\
    3D-GCK~\cite{gahlert2020single} & & & \cellcolor{gray!10} 5.79/3.27 & \cellcolor{gray!10} 4.57/2.52 & \cellcolor{gray!10} 3.64/2.11 \\
    \hline
    \hline
    WeakM3D~\cite{peng2022weakm3d} & \multirow{2} * {Weak} & Train &11.82/5.03 & 5.66/2.26 & 4.08/1.63 \\
    Our WeakMono3D & & None & \cellcolor{green!10} 12.31/6.98 & \cellcolor{green!10} 8.80/4.85 & \cellcolor{green!10} 7.81/4.45 \\
    \hline
    Improvement & - & - & +0.49/+1.95 & +3.14/+2.59 & +3.73/+2.82 \\
    \hline
    \end{tabular}}
  \caption{\textbf{Comparison on KITTI test set for the car category.} For all results, we use AP$|\scriptstyle R_{40}$ metrics with IoU threshold$=0.7$. The best results are highlighted in \textbf{bold} and our results are with \colorbox{green!10}{green} background. "Imporvement" means the improvement of our method compared with WeakM3D.}
  \label{tab:kitti_test}
   \vspace{-2mm}
\end{table*}

\subsection{Comparison on KITTI}
\noindent\textbf{KITTI}~\cite{kitti} is a popular 3D object detection benchmark that contains 7,481 images for training and 7,518 for testing. We follow the setting in~\cite{chen20153d} and split the original training set into 3,712 images for training and 3,769 images for validation. In this benchmark, there are three classes: Car, Pedestrian, and Cyclist. For each class, there are three levels of difficulty: Easy, Moderate, and Hard. Most weakly supervised methods in the 3D object detection area employ IOU 0.5 criterion on the valid set and IOU 0.7 criterion on the test set. In our experiments, we follow the same criterion as previous works. We evaluate our method on the validation set locally and submit the detection results to the official website to obtain the metric of the test set. 

To show the improvements, we compare our WeakMono3D with some representative monocular 3D object detection methods, including fully and weakly supervised ones. Table~\ref{tab:kitti_val} shows our comparison on the validation set. We achieved AP$_{3D}$ of $49.37,39.01,36.34$ and AP$_{BEV}$ of $54.32,42.83,40.07$, in terms of the Easy, Moderate and Hard categories. Compared with some weakly supervised methods, results show that even without needing LiDAR data, our method can achieve comparable performance with previous methods and significantly outperforms them in the Moderate and Hard categories. We attribute that 3D point clouds in the distance are sparse, leading to the previous models' performance degeneration. Images do not have this problem where all the objects have a dense representation, so our method performs well in the categories mainly containing far objects.
Moreover, our method performs comparably with some fully supervised methods without 3D annotation. In Table~\ref{tab:kitti_test}, we compare the metrics on the test set for cars with the prior methods. 
{This comparison is more convincing than the validation set because the ground truth labels of the test set are not publicly available.}
Our method obviously outperforms WeakM3D~\cite{peng2022weakm3d} on all the metrics, and performs comparably to some fully-supervised methods.
Table~\ref{tab:kitti_ped_cyc} shows our performance for pedestrian and cyclist class on the KITTI test set. Unlike some weakly supervised works~\cite{peng2022weakm3d,koestler2020learning} mainly focus on rigid objects like Cars, our method has better generality in pedestrians and cyclists, showing comparable performance with fully-supervised methods.

\noindent\textbf{Quantitive Results} Fig.~\ref{Fig:illustration} shows more visualizations results by our method. Owe to guidance from the projection loss, the visualized 3D boxes in the 2D images fit well with the corresponding targets. {ccording to the BEV illustration, the predictions have achieved accurate depth values and rotation angles, since the consistency loss provides a comprehensive 3D spatial guidance and the rotation loss corrects the direction difference between 2D ground truth and predictions.}

\begin{table*}
  \small
  \centering
    \setlength{\tabcolsep}{5mm}{
    \begin{tabular}{l|ccc|ccc} 
    \hline
    \multirow{2}*{Method}  & \multicolumn{3}{c|}{Pedestrian} & \multicolumn{3}{c}{Cyclist} \\
     \cline{2-4}\cline{5-7} & Easy & Moderate & Hard & Easy & Moderate & Hard \\
    \hline
    OFTNet~\cite{roddick2018orthographic} & 1.28/0.63 & 0.81/0.36 & 0.51/0.35 & 0.36/0.36 & 0.16/0.16 & 0.15/0.15 \\
    \rowcolor{gray!10} SSD3D~\cite{jorgensen2019monocular} & 2.48/2.31 & 2.09/1.78 & 1.61/1.48 & 3.45/- & 1.89/- & 1.44/- \\
    M3D-RPN~\cite{brazil2019m3d} & 5.65/4.92 & 4.05/3.48 & 3.29/2.94 & 1.25/0.94 & 0.81/0.65 & 0.78/0.47 \\
    \rowcolor{gray!10} MonoPSR~\cite{ku2019monocular} & 7.24/6.12 & 4.56/4.00 & 4.11/3.30 & \textbf{9.87/8.47} & \textbf{5.78/4.74} & \textbf{4.57/3.68} \\
    MonoDis~\cite{simonelli2020disentangling} & 9.07/7.79 & 5.81/5.14 & 5.09/4.42 & 1.47/1.17 & 0.85/0.54 & 0.61/0.48 \\
    \rowcolor{gray!10} DD3D~\cite{dd3d} & \textbf{15.90/13.91} & \textbf{10.85/9.30} & \textbf{8.05/8.05} & 3.20/2.39 & 1.99/1.52 & 1.79/1.31 \\
    \hline
    \hline
    \rowcolor{green!10} Ours & 5.53/4.62 & 3.86/3.13 & 3.29/2.87 & 3.04/2.25 & 1.87/1.41 & 1.59/1.22 \\
    \hline
    \end{tabular}}
  \caption{\textbf{Comparison on KITTI test set for Pedestrian and Cyclist category.} For all results, we use AP$|\scriptstyle R_{40}$ metrics with IoU threshold$=0.7$. We highlight the best results in \textbf{bold} and our results in \colorbox{green!10}{green}.}
  \label{tab:kitti_ped_cyc}
  \vspace{-2mm}
\end{table*}

\begin{table}
  \small
  \centering
    \setlength{\tabcolsep}{1mm}{
    \begin{tabular}{cc|cc}
    \hline
    \multicolumn{2}{c|}{AP$_{BEV}$/AP$_{3D}$ (IoU=0.5)$|\scriptstyle R_{40}$} &
    \multicolumn{2}{c}{AP$_{BEV}$/AP$_{3D}$ (IoU=0.7)$|\scriptstyle R_{40}$} \\
    \hline
    Easy & Hard & Easy & Hard \\
    \hline
    34.85/32.80 & 24.03/22.07 & 17.58/14.30 & 10.37/7.32 \\
    \hline
    \end{tabular}}
  \caption{\textbf{Results on the proposed ProdCars dataset.}}
  \label{tab:ablation_prodcars}
  \vspace{-2mm}
\end{table}

\begin{table}
  \small
  \centering
    \setlength{\tabcolsep}{1mm}{
    \begin{tabular}{c|c|c|c|ccc}
    \hline
    \multirow{2}*{\#} & 
    \multirow{2}*{${L}_{proj}$} &
    \multirow{2}*{${L}_{con}$} &  \multirow{2}*{${L}_{rot}$} & 
    \multicolumn{3}{c}{AP$_{BEV}$/AP$_{3D}$ (IoU=0.5)$|\scriptstyle R_{40}$} \\
    \cline{5-7}
    &  & & & Easy & Moderate & Hard \\
    \hline
     0 & \cmark & & & 0.00/0.00 & 0.00/0.00 & 0.00/0.00 \\
    \rowcolor{gray!10} 1 & \cmark  & \cmark & & 17.36/11.78 & 15.27/9.58 & 14.23/9.13 \\
   
    2 &  \cmark& \cmark & \cmark& \textbf{54.32/49.37} & \textbf{42.83/39.01} & \textbf{40.07/36.34} \\
    \hline
    \end{tabular}}
  \caption{\textbf{Ablation study for different losses.}}
  \label{tab:ablation_loss}
   \vspace{-3mm}
\end{table}

\begin{table}
  \small
  \centering
  \setlength{\tabcolsep}{1.5mm}{
    \begin{tabular}{c|c|ccc}
    \hline
    \multirow{2}*{\#} & 
    \multirow{2}*{Source} & 
    \multicolumn{3}{c}{AP$_{BEV}$/AP$_{3D}$ (IoU=0.5)$|\scriptstyle R_{40}$} \\
    \cline{3-5} & & Easy & Moderate & Hard \\
    \hline
    0 & Multi-Camera & \textbf{54.32/49.37} & \textbf{42.83/39.01} & \textbf{40.07/36.34} \\
    1 & Sequence & 43.04/36.68 & 31.57/26.80 & 27.18/23.33 \\
    \hline
    \end{tabular}}
  \caption{\textbf{Ablation study for the multi-view data sources.}}
  \label{tab:ablation_policy}
   \vspace{-2mm}
\end{table}

\begin{table}
  \small
  \centering
    \setlength{\tabcolsep}{1.5mm}{
    \begin{tabular}{c|c|ccc}
    \hline
    \multirow{2}*{\#} & 
    \multirow{2}*{Ratio} & 
    \multicolumn{3}{c}{AP$_{BEV}$/AP$_{3D}$ (IoU=0.5)$|\scriptstyle R_{40}$} \\
    \cline{3-5} & & Easy & Moderate & Hard \\
    \hline

    DD3D & $1$ & 64.09/58.08 & 51.84/46.78 & 47.52/42.70 \\
    \hline
    \hline
    0 & $0$ & 54.32/49.37 & 42.83/39.01 & 40.07/36.34 \\
    \rowcolor{gray!10} 1 & $1/20$ & 59.47/53.61 & 48.11/43.72 & 43.27/39.45 \\
    2 & $1/10$ & 63.72/57.21 & 49.95/44.60 & 44.54/40.50 \\
    \rowcolor{gray!10} 3 & $1/5$ & 63.92/59.12 & 50.52/46.46 & 46.29/41.66 \\
    4 & $1/3$ & \textbf{66.36/61.02}& \textbf{53.00/47.69} & \textbf{48.48/43.56} \\
    \hline
    \end{tabular}}
  \caption{\textbf{Ablation study of integration with the fully supervised method.}}
  \label{tab:ablation_ratio}
  \vspace{-4mm}
\end{table}

\subsection{ProdCars Dataset}
To prove that our method can work well with {the production data}, we collect a new dataset from production cars and name it as {\textbf{ProdCars}}. There are about 10,000 image frames in the training set, each containing 2D boxes and 2D direction labels without point clouds. We also collect a new validation set to evaluate the performance. The validation set is collected by another car with the same cameras as production cars with one LiDAR sensor. Therefore, these images can be labeled with 3D LiDAR ground truths for evaluation purposes. Table~\ref{tab:ablation_prodcars} show the results of our WeakMono3D on ProdCars.

\subsection{Ablation Study}
To verify the significance of the proposed method, we conduct several ablation studies on KITTI validation dataset. We mainly focus on three aspects. The first one is the three weakly supervised losses proposed in this paper. Then the second one is the integration with the fully supervised method. And finally, we discuss the influence of different types of multi-view data.

\noindent\textbf{Losses.} In this paper, we discuss three types of consistency and propose three corresponding loss functions, $L_{proj}$, $L_{con}$ and $L_{rot}$. Table~\ref{tab:ablation_loss} shows their ablation studies of them. The model with only projection loss shows a totally zero result because the projection loss itself cannot correct the prediction error in the 3D space. Therefore the final predictions have accurate projections in the image but wrong positions in the real world, producing zero metrics. Direction consistency loss is also critical to the performance of models. The optimization of size prediction relies on minimizing the projection error. If the rotation of objects is not accurate, this loss will guide the model to provide the wrong sizes to fit the 2D ground truth, leading to a bad performance. In short, the three losses are all indispensable in our weakly supervised method. Only by employing them together can we give comprehensive guidance for the optimization of the detection models.

\noindent\textbf{{Employed as} Pre-training Method.} One of the main purposes of this paper is to enhance the detection model by exploiting the feedback data. 
To show the feasibility of this, we employ our method as a pre-training manner and fine-tune the model with a small proportion of fully labeled data. 
As shown in Table~\ref{tab:ablation_ratio}, even fine-tuned with an extremely small amount of data, our model can achieve comparable performance with the corresponding fully supervised model. With about 1/3 of fully labeled data, our method can significantly outperform the fully supervised baseline DD3D~\cite{dd3d}.
 
\noindent\textbf{Different Types of Multi-view Data.}
As we discussed in \S~\ref{Sec:consistency_loss}, multi-view images can be taken from different sources of data. Table~\ref{tab:ablation_policy} shows the comparison of two sources, the multi-camera data, or the video sequences from one single camera. 
We employ the left and right cameras in the KITTI dataset as multi-camera data, 
and frames from the left camera as the video sequences in our experiments. Experiments show that multi-camera data can achieve better performances than sequence data. In other experiments on KITTI dataset, we only report the results of the multi-camera trained model.

\section{Conclusion}
In this paper, we have proposed a weakly supervised method that can train the monocular 3D object detection models with only 2D labels marked on images. 
{To achieve this, we explore three types of consistency, \ie projection, multi-view, and direction consistency, and design three losses based on them.
To show the feasibility, we propose a new dataset {\textbf{ProdCars}} collected by the production cars, and experiments show that our method can work well on this data.
{When employed as a pre-training method}, our model can significantly outperform the corresponding fully supervised version with only 1/3 of labeled data. 
Experiments show that our trained models can achieve comparable performance with some fully supervised methods. 
We believe the proposed method can help to use the rich data from the production cars, because these data contain a larger variety of scenes than experimentally collected data, which is critical to improve the robustness and generality of models. 
}

{\small
\bibliographystyle{ieee_fullname}
\bibliography{egbib}
}

\end{document}